\documentclass{llncs}

\newcommand{\ignore}[1]{}

\def\ss{\smallskip}

\usepackage{latexsym}
\usepackage{amsmath,amssymb}
\usepackage[decmulti]{inputenc}
\usepackage{verbatim}
\usepackage{graphicx}

\begin{document}
\title{{\sc Physical Simulation of Inarticulate Robots}}

\author{%
}

\institute{%
{\normalsize
\begin{tabular}{c}
{\bf Guillaume Claret}, {\bf Micha\"{e}l Mathieu}, {\bf David Naccache} and {\bf Guillaume Seguin}\\
\'Ecole normale sup\'erieure\\
D\'epartement d'informatique\\
45 rue d'Ulm, {\sc f}-75230, Paris {\sc cedex 05}, France\\
{\tt \{surname.name\}@ens.fr} except {\tt mmathieu@clipper.ens.fr}\\
\end{tabular}
}}
\maketitle

\begin{abstract}

In this note we study the structure and the behavior of inarticulate robots. We introduce a robot that moves by successive revolvings. The robot's structure is analyzed, simulated and discussed in detail.

\end{abstract}

\section{Introduction}

In this note we study the structure and the behavior of inarticulate robots. The rationale for the present study is the fact that, in most robots,
articulations are one of the most fragile system parts. Articulations require lubricants and call for regular maintenance which might be impossible in radioactive, subaquatic or space environments. In addition, articulations are sensitive to dust (or humidity) and must hence be shielded from external nano-particles {\sl e.g.} during martian sand-storms.\smallskip

In this work we circumvent articulations by studying a robot that moves by shifting its center of gravity so as to flip repeatedly.

\section{The Robot}

\begin{figure}[h]
         \begin {center}
\includegraphics [height = 5cm]{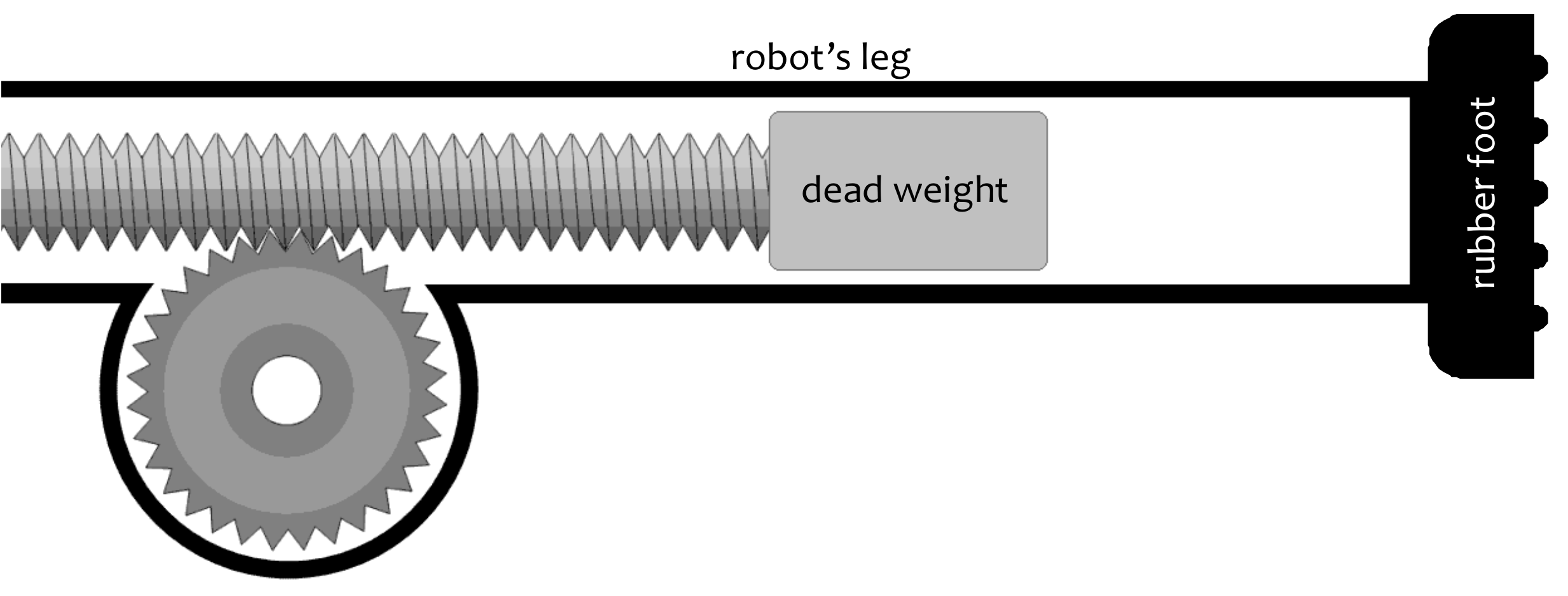}
\caption {Schematic Cross-Section of the Robot's Leg}
\label{leg}
            \end {center}
        \end {figure}

The proposed robot's model is a regular polyhedron prolonged with hollow legs. Each hollow leg contains a worm drive allowing to move an internal mass $m$ inside the leg\footnote{The internal mass must not necessarily be a dead weight. {\sl e.g} it can be the battery used to power the worm drive.} as shown in Figure \ref{leg}. By properly moving the masses the device manages to revolve and hence move in the field.\smallskip

\begin{figure}[h]
         \begin {center}
\includegraphics [height = 5cm]{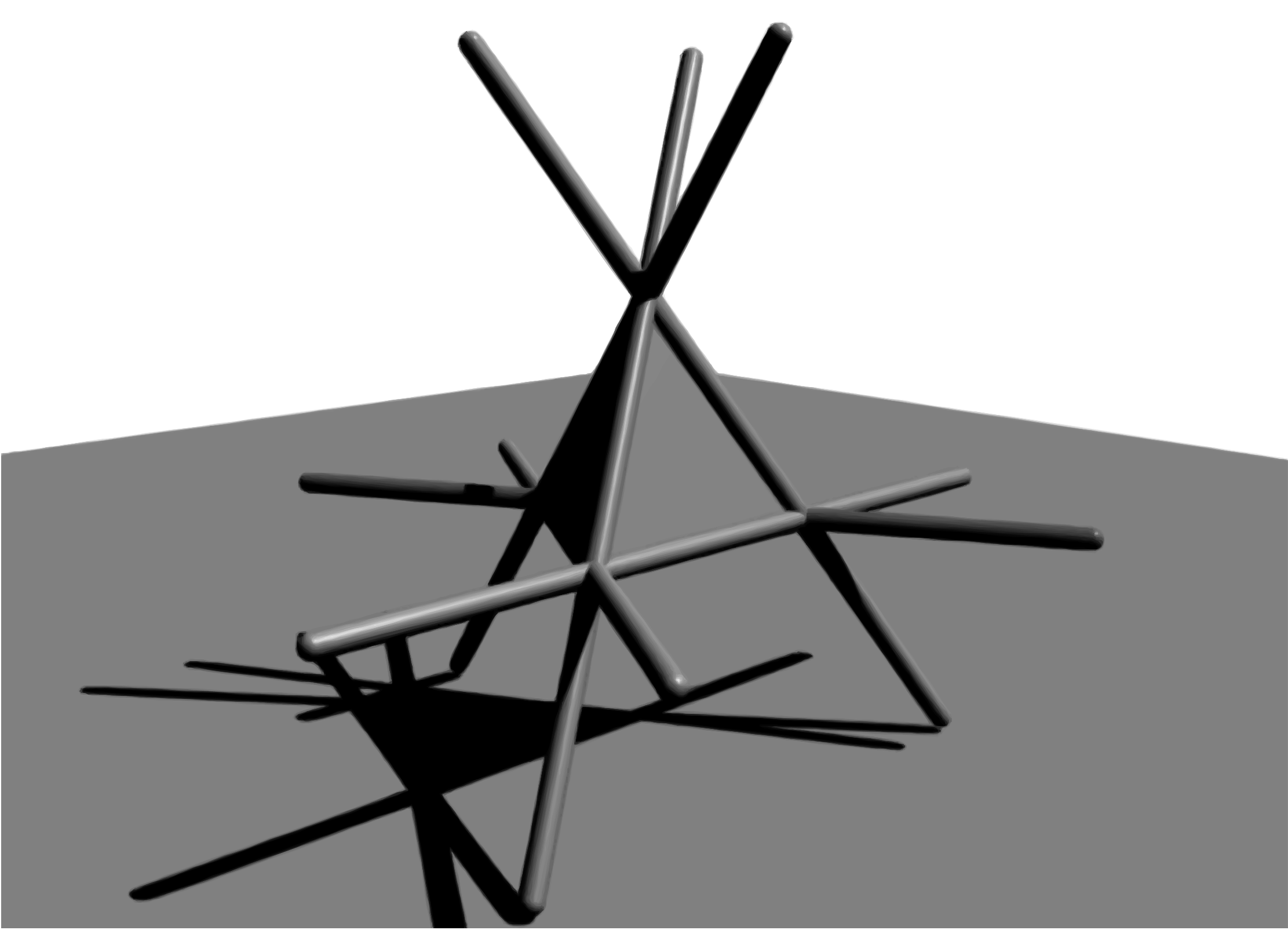}
\caption {Basic Robot Structure, Head-Up ({\sc hu}) State.}
\label{hu}
            \end {center}
        \end {figure}

\begin{figure}[h]
         \begin {center}
\includegraphics [height = 5cm]{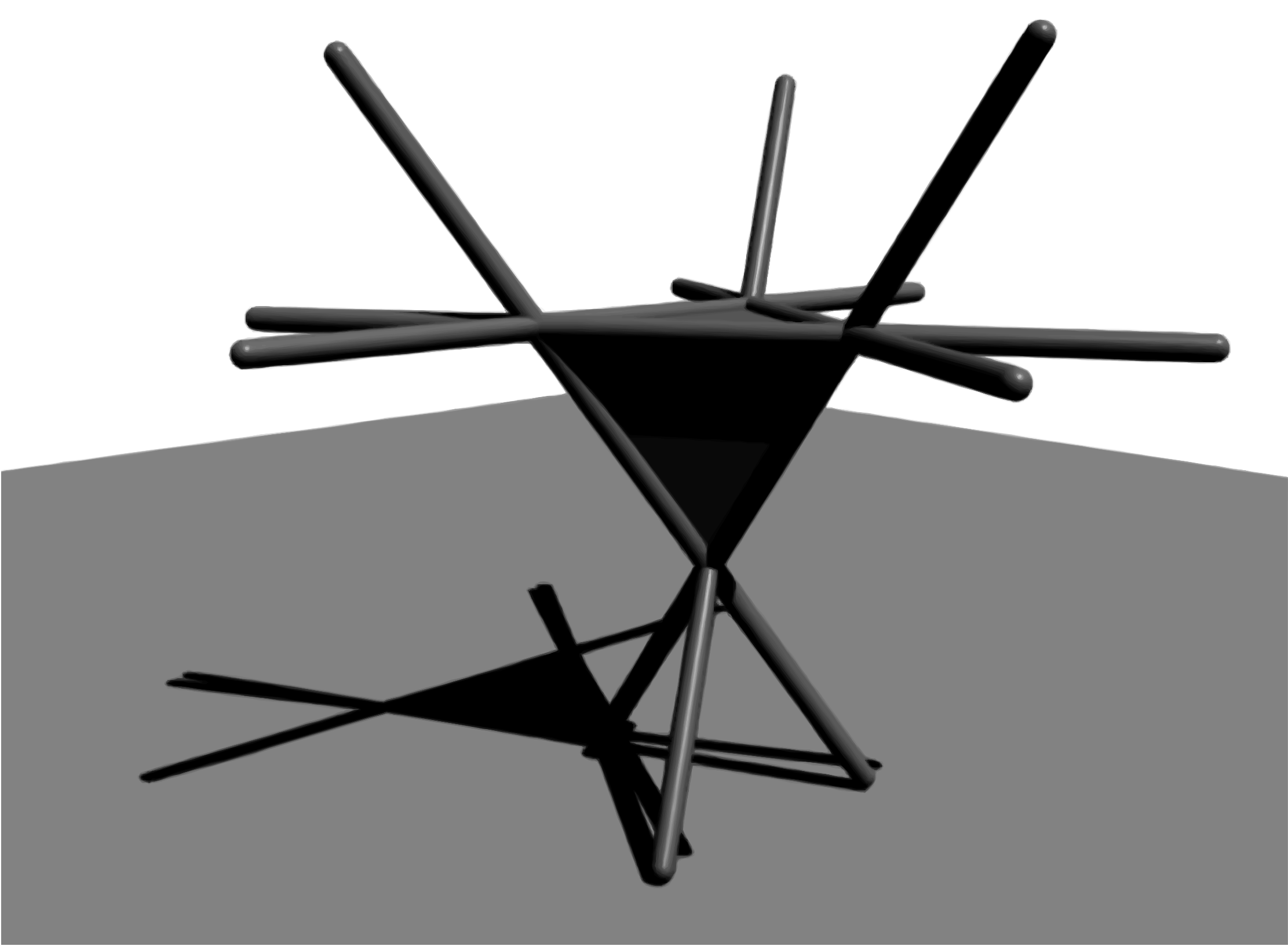}
\caption {Basic Robot Structure, Head-Down ({\sc hd}) State.}
\label{hd}
            \end {center}
        \end {figure}

\begin{figure}[h]
         \begin {center}
\includegraphics [height = 5cm]{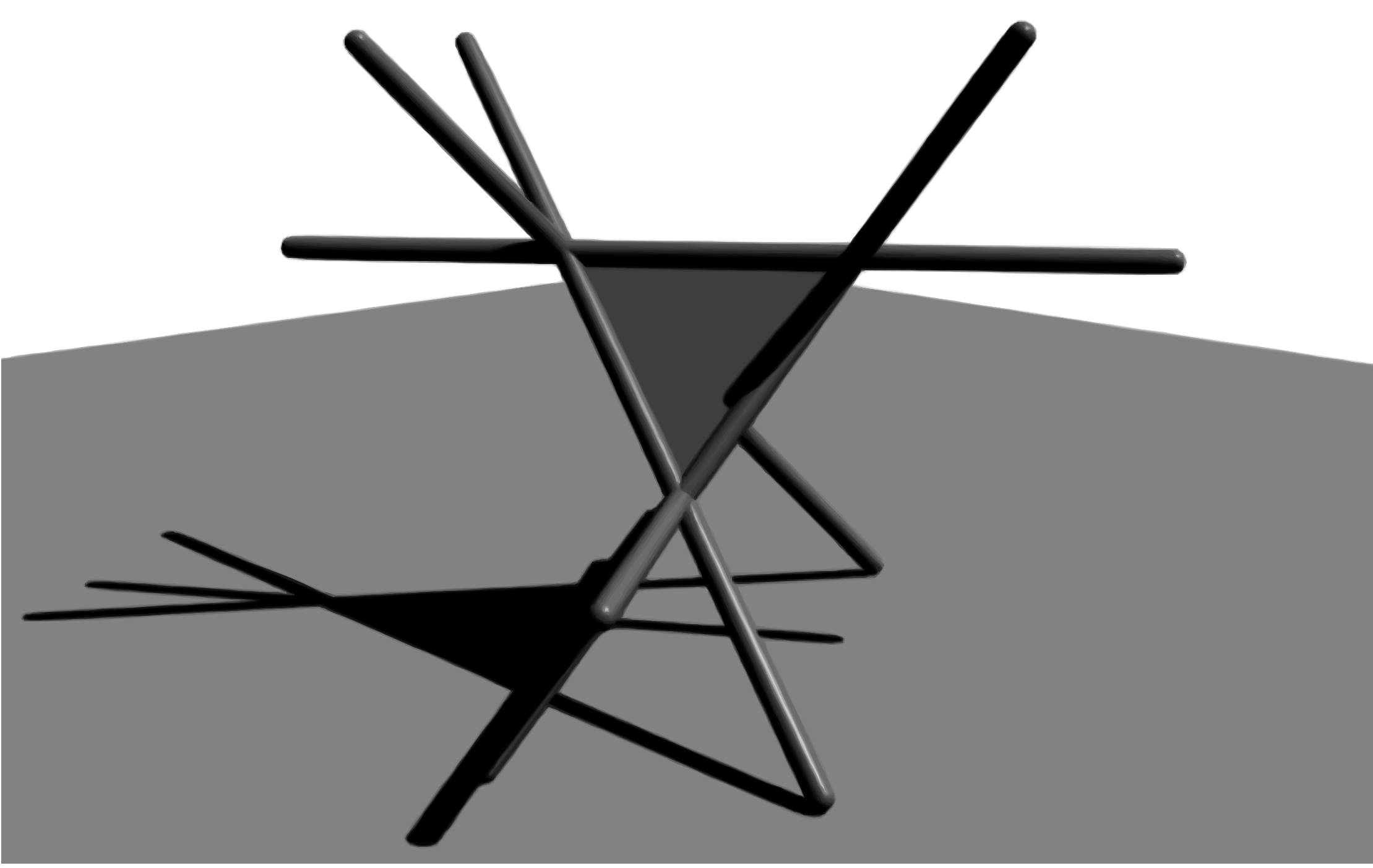}
\caption {Basic Robot Structure, Side-Down ({\sc sd}) State.}
\label{sd}
            \end {center}
        \end {figure}

Different regular polyhedra can be used as robot bodies. In this study we chose the simplest, namely a tetrahedron. Hence, the robot has two basic geometrical parameters, $\ell$ the tetrahedron's edge and $L$ the leg's length. Figures \ref{hu}, \ref{hd} and \ref{sd} show the robot's structure.\smallskip

The robot has three stable states, head-down ({\sc hd}), head-up ({\sc hu}) and side-down ({\sc sd}). In the head-down and head-up states, the robot rests on three legs while in the side-down mode the robot rests on four legs. Possible transition modes are hence:

$$
\mbox{head-down}  \leftrightarrow  \mbox{side-down} \leftrightarrow  \mbox{head-up}
$$

Note that a direct $\mbox{head-down}  \leftrightarrow  \mbox{head-up}$ transition is impossible.\smallskip

\begin{figure}[h]
         \begin {center}
\includegraphics [height = 9cm]{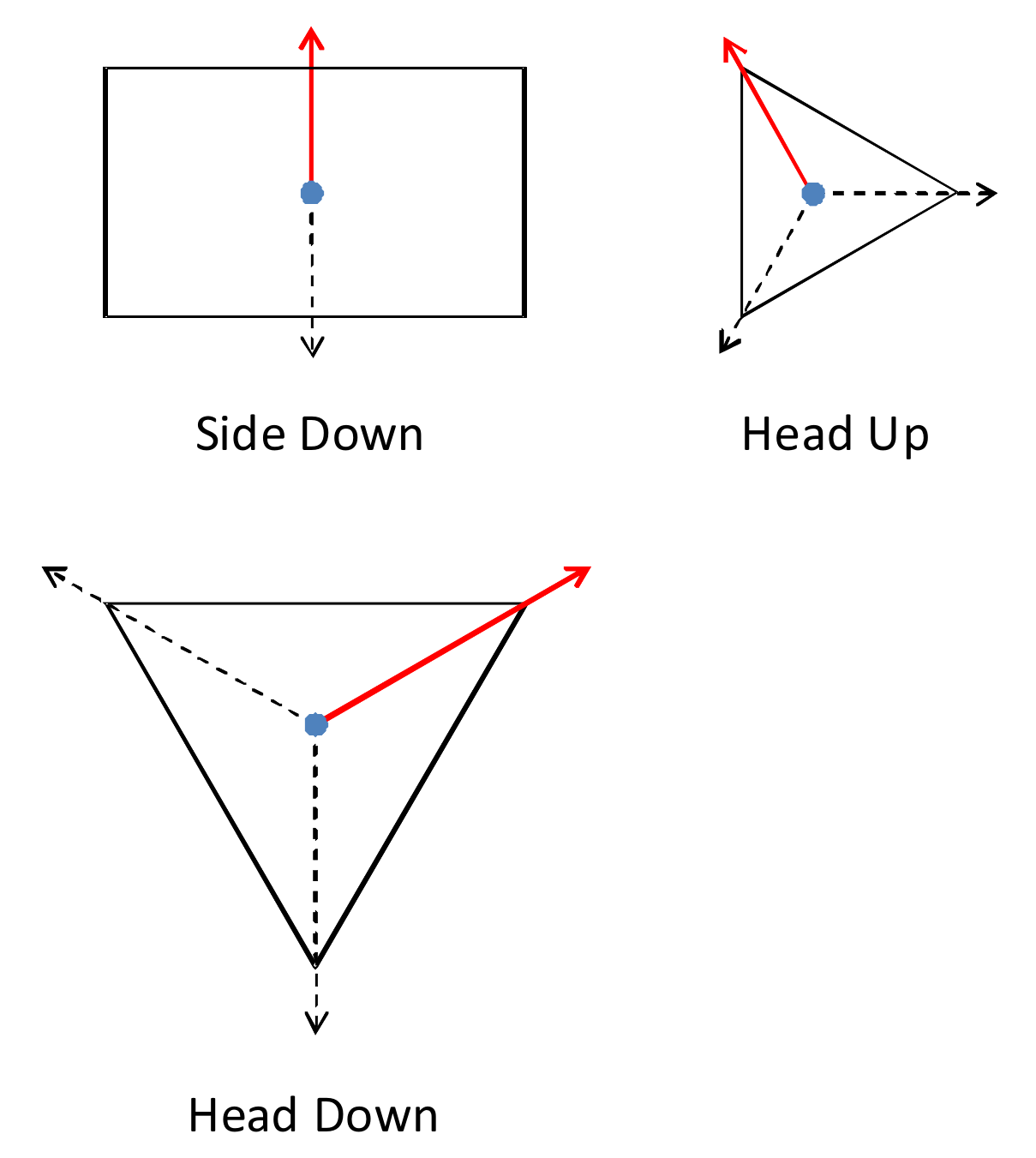}
\caption {Reference Directions}
\label{faces}
            \end {center}
        \end {figure}

The robot's state and position are thoroughly characterized by three parameters: $G=\{G_X,G_Y\}$ the $\{X,Y\}$ coordinates of the robot's centroid,
$P\in\{\mbox{{\sc hd}},\mbox{{\sc hu}},\mbox{{\sc sd}}\}$ the robot's current stable state and the angle $\alpha$ formed between the $X$ axis and the robot's reference direction. The reference direction, shown in Figure \ref{faces}, is defined in two different ways depending on the robot's current state.\ss

\section{Reachable Points}

\begin{figure}[h]
         \begin {center}
\includegraphics [height = 6cm]{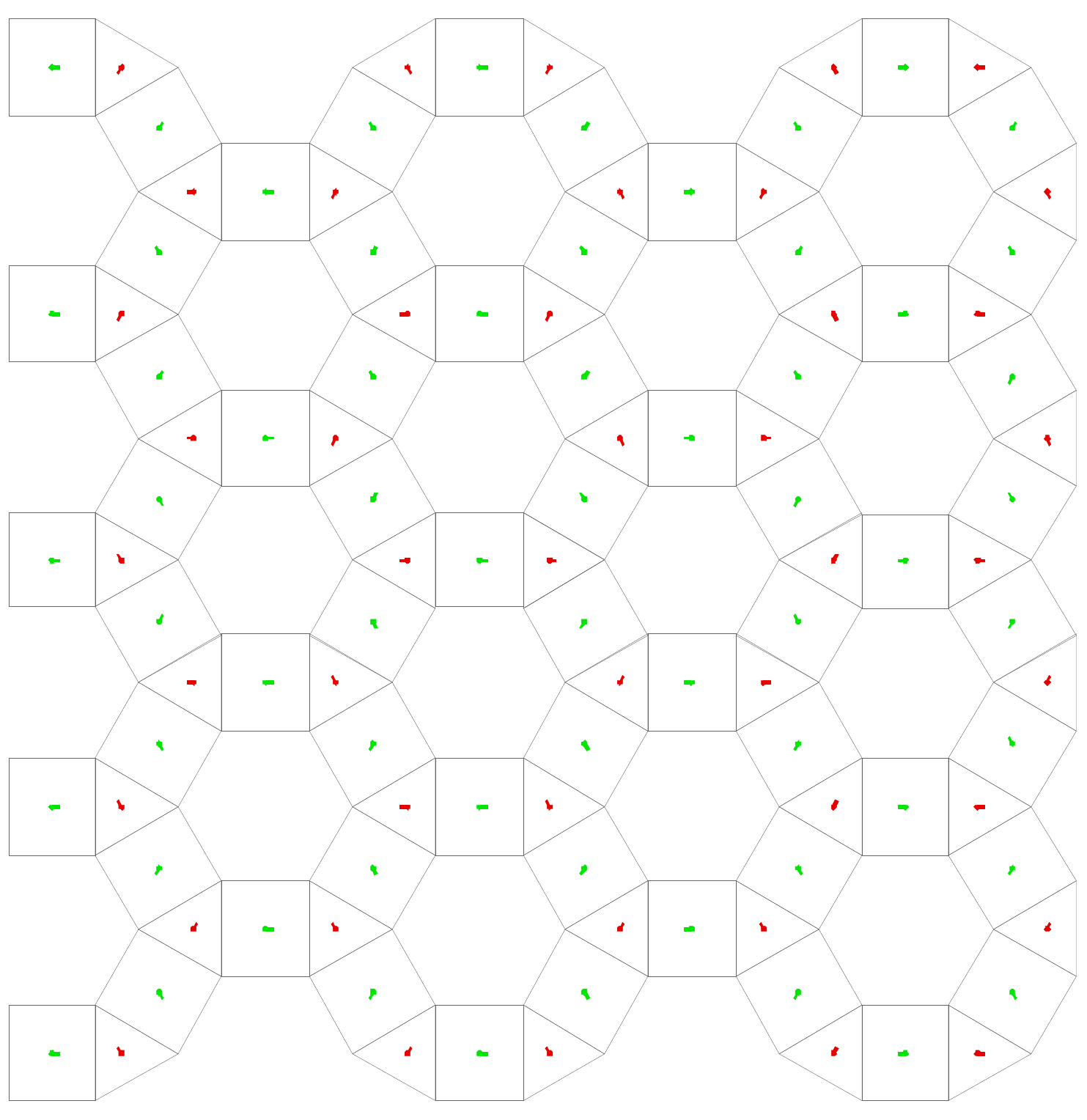}
\caption {Bistate Locmotion $\mbox{{\sc sd}}\leftrightarrow\mbox{{\sc hu}}$}
\label{bistate1}
            \end {center}
        \end {figure}

\begin{figure}[h]
         \begin {center}
\includegraphics [height = 6cm]{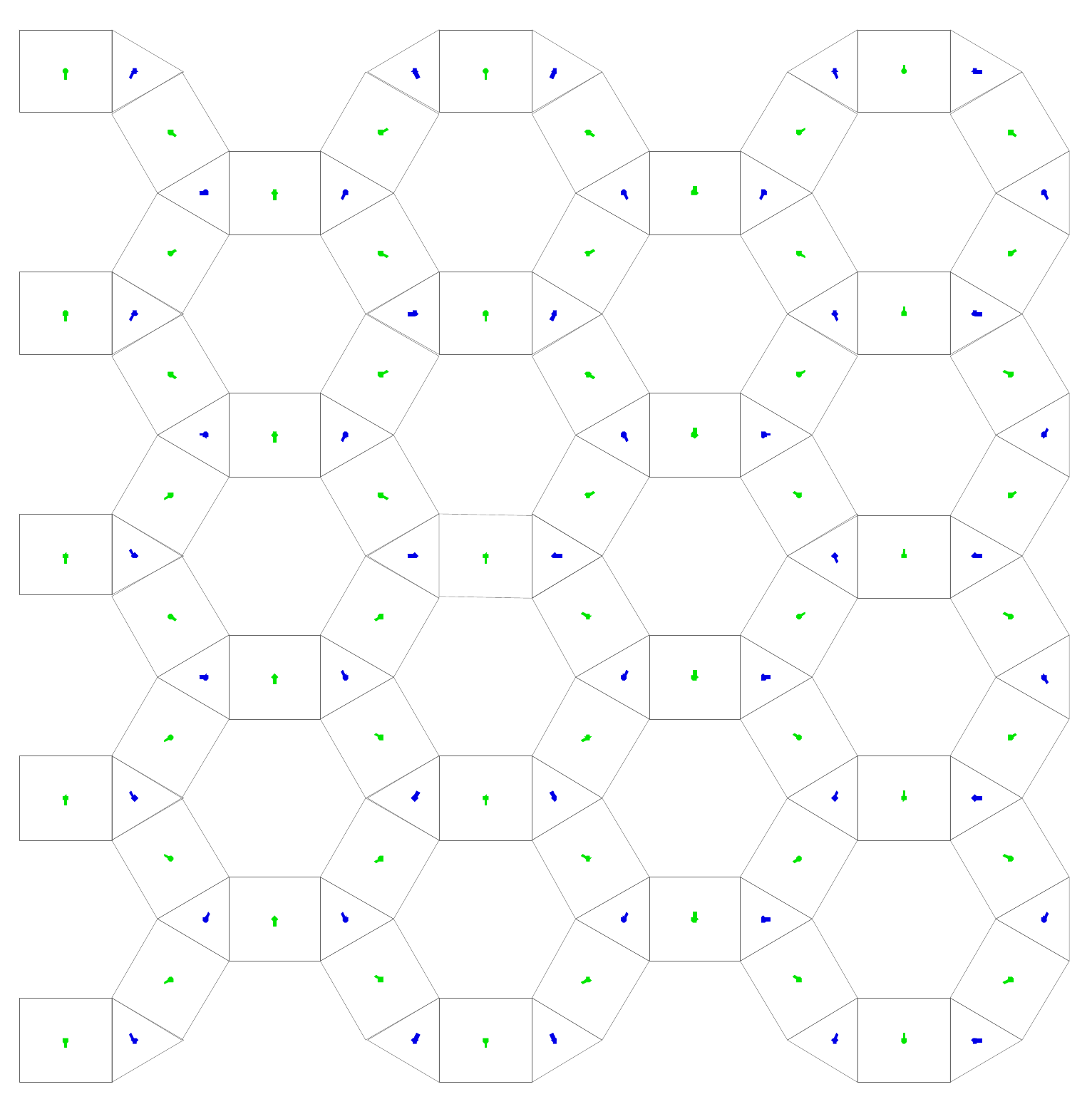}
\caption {Bistate Locmotion $\mbox{{\sc hd}}\leftrightarrow\mbox{{\sc sd}}$}
\label{bistate2}
            \end {center}
        \end {figure}

\begin{figure}[h]
         \begin {center}
\includegraphics [height = 6cm]{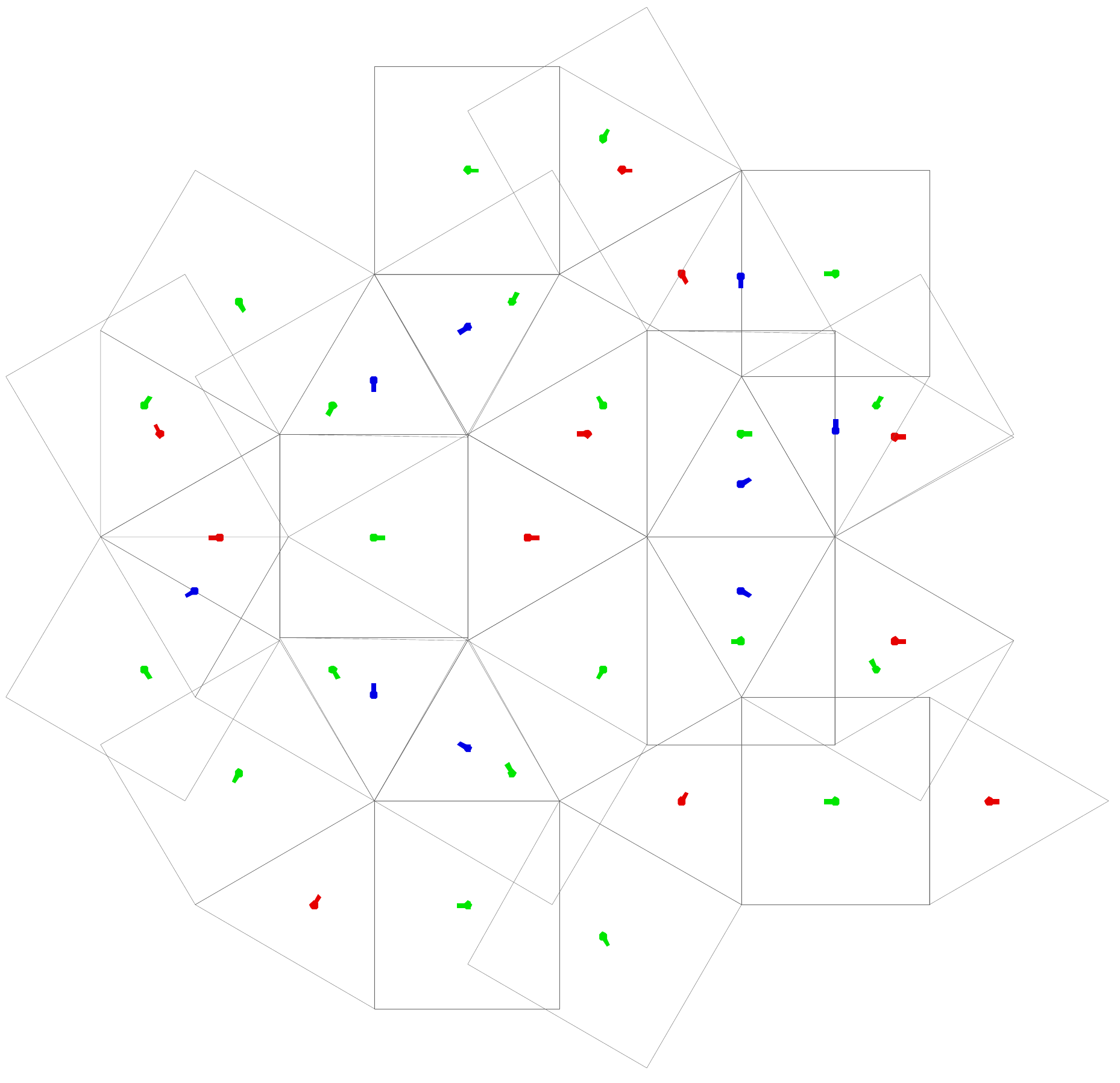}
\caption {Tristate Locmotion $\mbox{{\sc hd}}\leftrightarrow\mbox{{\sc sd}}\leftrightarrow\mbox{{\sc hu}}$}
\label{tristate}
            \end {center}
        \end {figure}

We define a {\sl reachable point} as any space coordinate on which the robot can set the center of the rubber foot. It appears (although we did not prove this formally) that when the robot is constrained to a bi-state ({\sl i.e.} $\mbox{{\sc hd}}\leftrightarrow\mbox{{\sc sd}}$ or $\mbox{{\sc sd}}\leftrightarrow\mbox{{\sc hu}}$) locomotion mode and to a delimited planar surface only a finite number of points can be reached (Figures \ref{bistate1} and \ref{bistate2}) whereas if we allow tristate $\mbox{{\sc hd}}\leftrightarrow\mbox{{\sc sd}}\leftrightarrow\mbox{{\sc hu}}$ transitions, an infinity of points seems to become reachable (Figures \ref{tristate} and \ref{maillage}). It might be the case that increasing the set of reachable points calls for walking further and further away from the robot's departure point and heading back to the vicinity of the departure point through a different path. Proving that an infinity of reachable points can be achieved in a delimited planar surface is an open question.\smallskip

\section{Pathfinding}

\begin{figure}[h]
         \begin {center}
\includegraphics [height = 8cm]{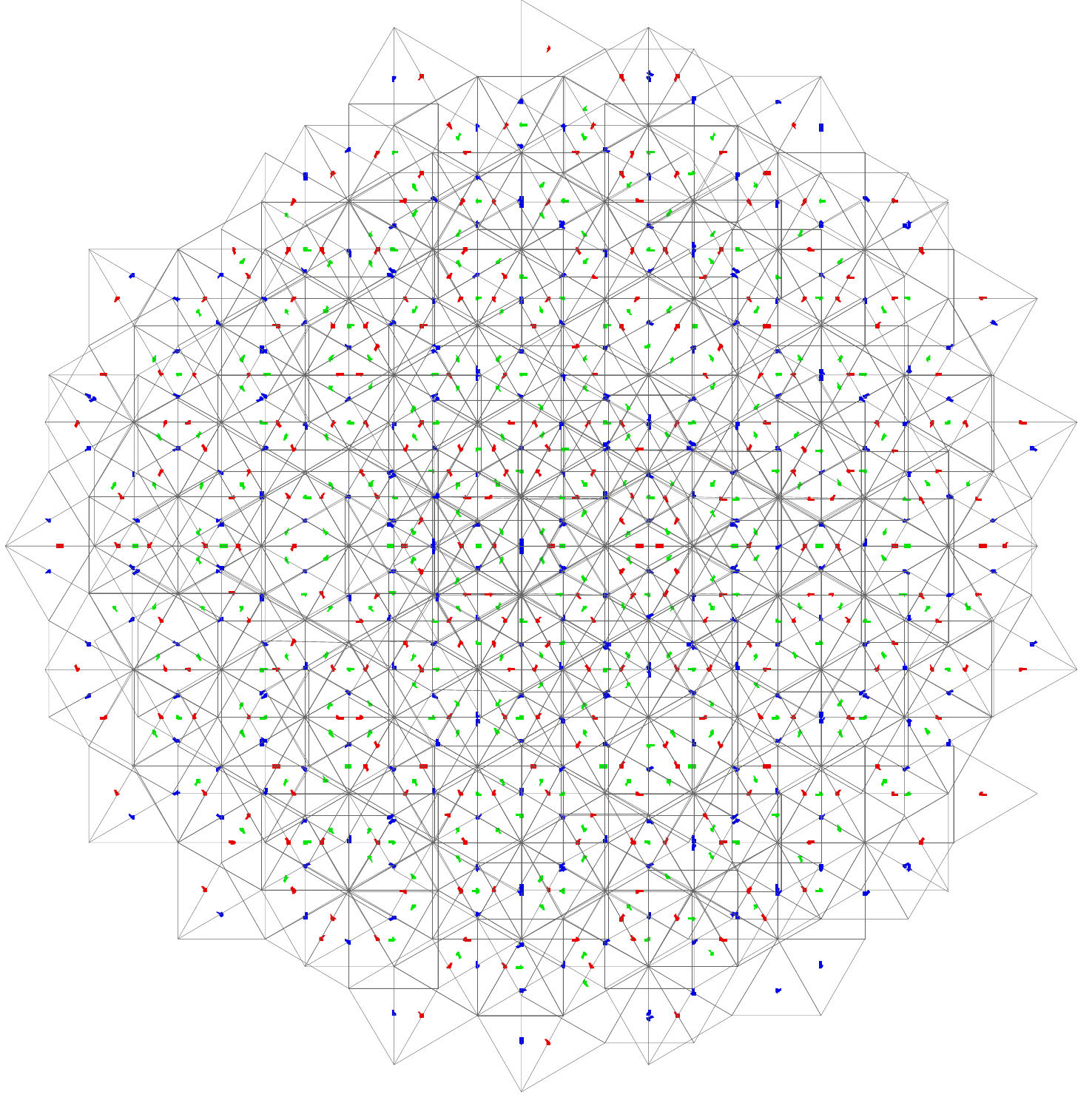}
\caption {Tristate Breadth First Search}
\label{maillage}
            \end {center}
        \end {figure}

To approximately reach a destination point, we first experimented a simple {\sc bfs} (Breadth First Search) algorithm \cite{bfs}. Before queuing potential revolving options, our implementation checked that the targeted position does not fall within an obstacle. This allowed locomotion with obstacle avoidance. The approach turned-out to be inefficient. Indeed, the $\mbox{{\sc hd}}\leftrightarrow\mbox{{\sc sd}}\leftrightarrow\mbox{{\sc hu}}$ locomotion results in the re-exploration of the already visited areas even though the algorithm records all already visited configurations. This typically happens when the edge of a rectangle and the edge of a triangle nearly overlap ({\sl cf.} Figure \ref{maillage}).\smallskip

To improve performance we implemented an $A^*$ algorithm \cite{astar}. This was done by modifying the {\sc bfs} simple queue into a prioritized queue. Priorities were determined using $\Delta_{\mbox{{\scriptsize dep}}}$, the length of the path since the departure point and an estimate of the distance to destination $\Delta_{\mbox{{\scriptsize des}}}$. At any step, the next chosen path is the shortest, {\sl i.e.} the one whose $\Delta_{\mbox{{\scriptsize des}}}+\Delta_{\mbox{{\scriptsize dep}}}$ is the smallest.\smallskip

\begin{figure}[h]
         \begin {center}
\includegraphics [height = 8cm]{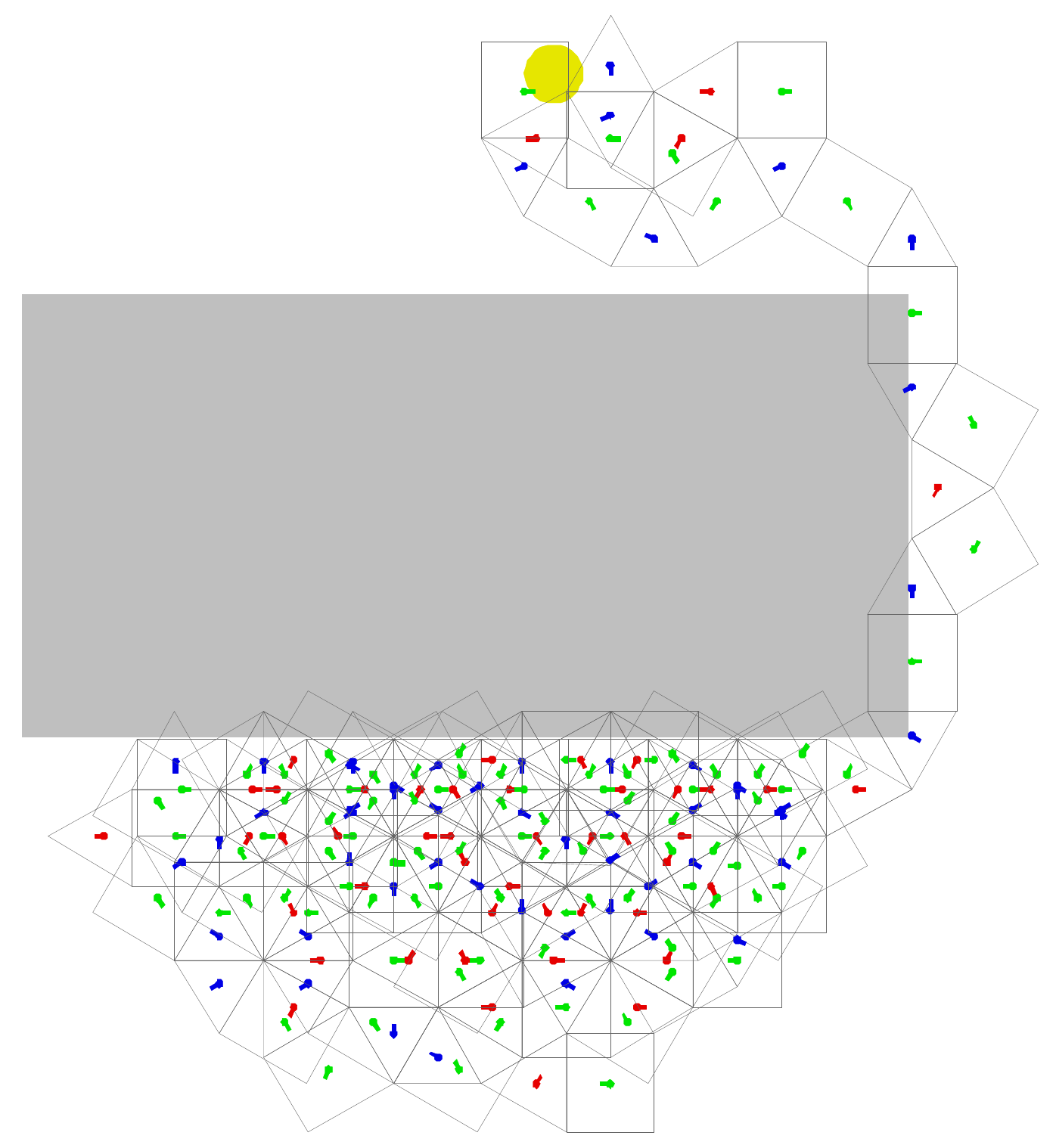}
\caption {$A^*$ Obstacle Avoidance}
\label{astar}
            \end {center}
        \end {figure}

The application of the $A^*$ algorithm to obstacle avoidance is depicted in Figure \ref{astar}. The yellow circle represents the arrival's target and the black rectangle is an obstacle. The obstacle avoidance C++ code can be downloaded from \cite{claret}.

\section{Simulation}

\begin{figure}[h]
         \begin {center}
\includegraphics [height = 7cm]{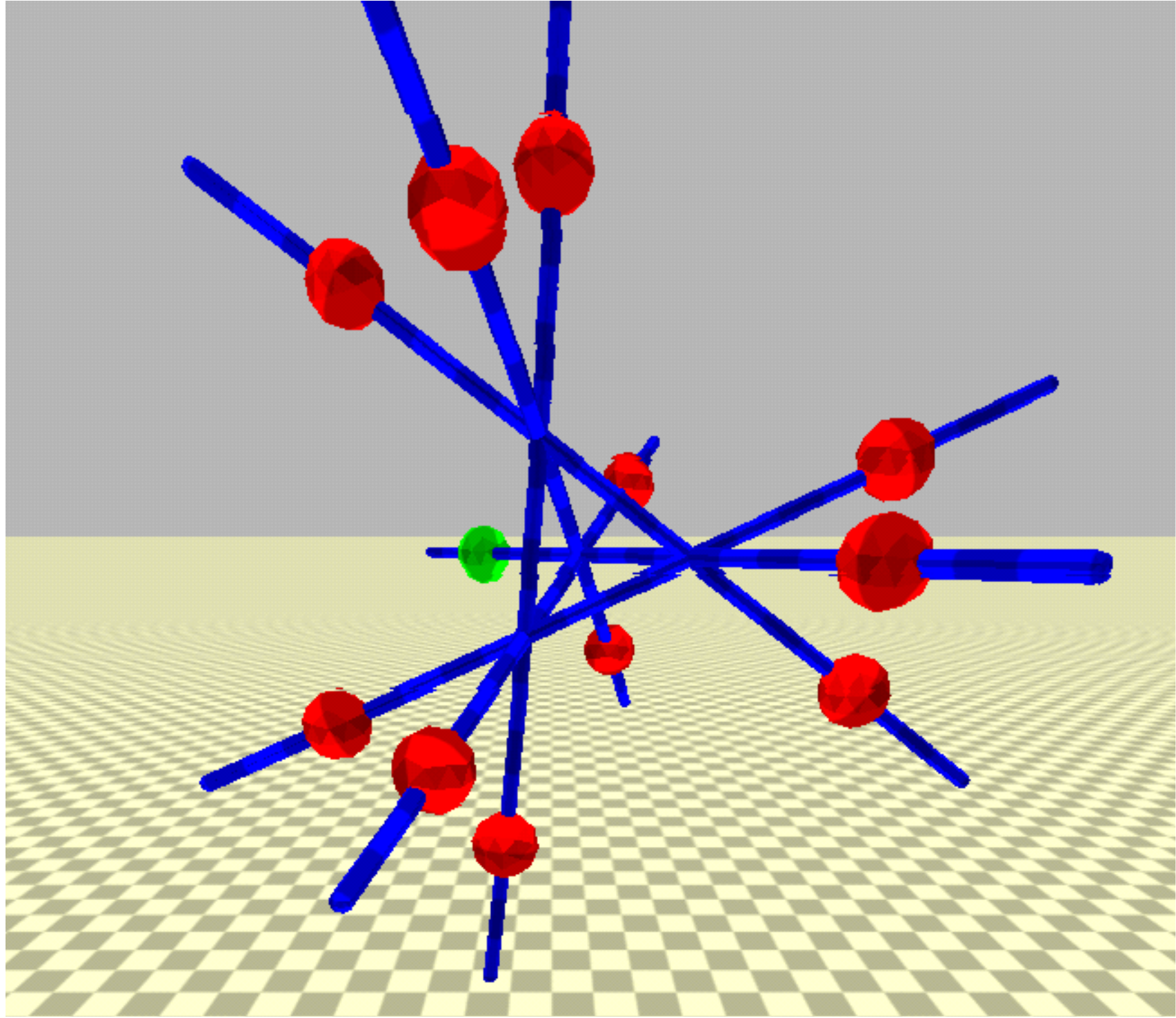}
\caption {Bullet Simulation, Details of The Robot}
\label{closeup}
            \end {center}
        \end {figure}

\begin{figure}[h]
         \begin {center}
\includegraphics [height = 7cm]{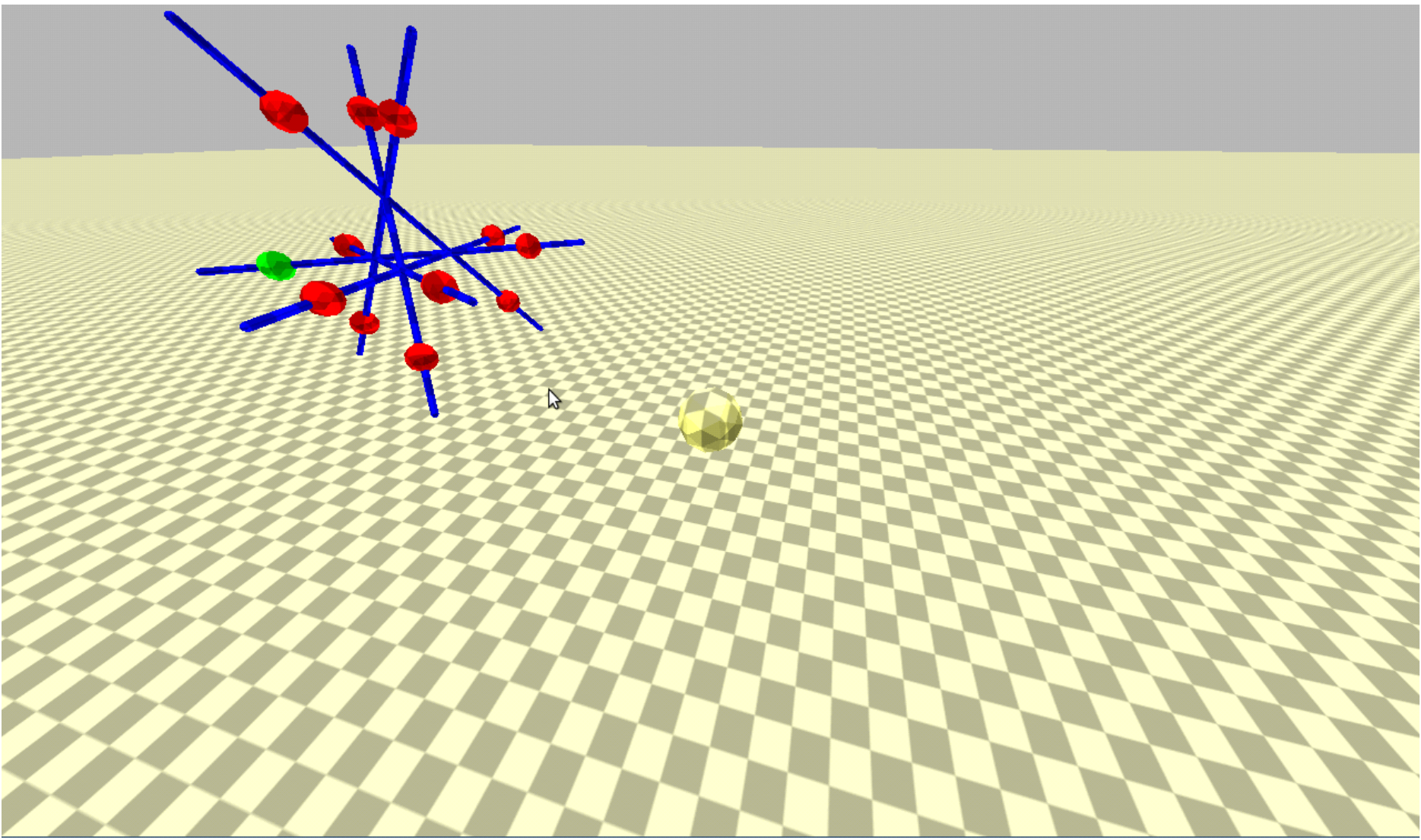}\\[0.8cm]
\caption {Bullet Simulation - Planar Locomotion}
\label{planar}
            \end {center}
        \end {figure}

\begin{figure}[h]
         \begin {center}
\includegraphics [height = 7cm]{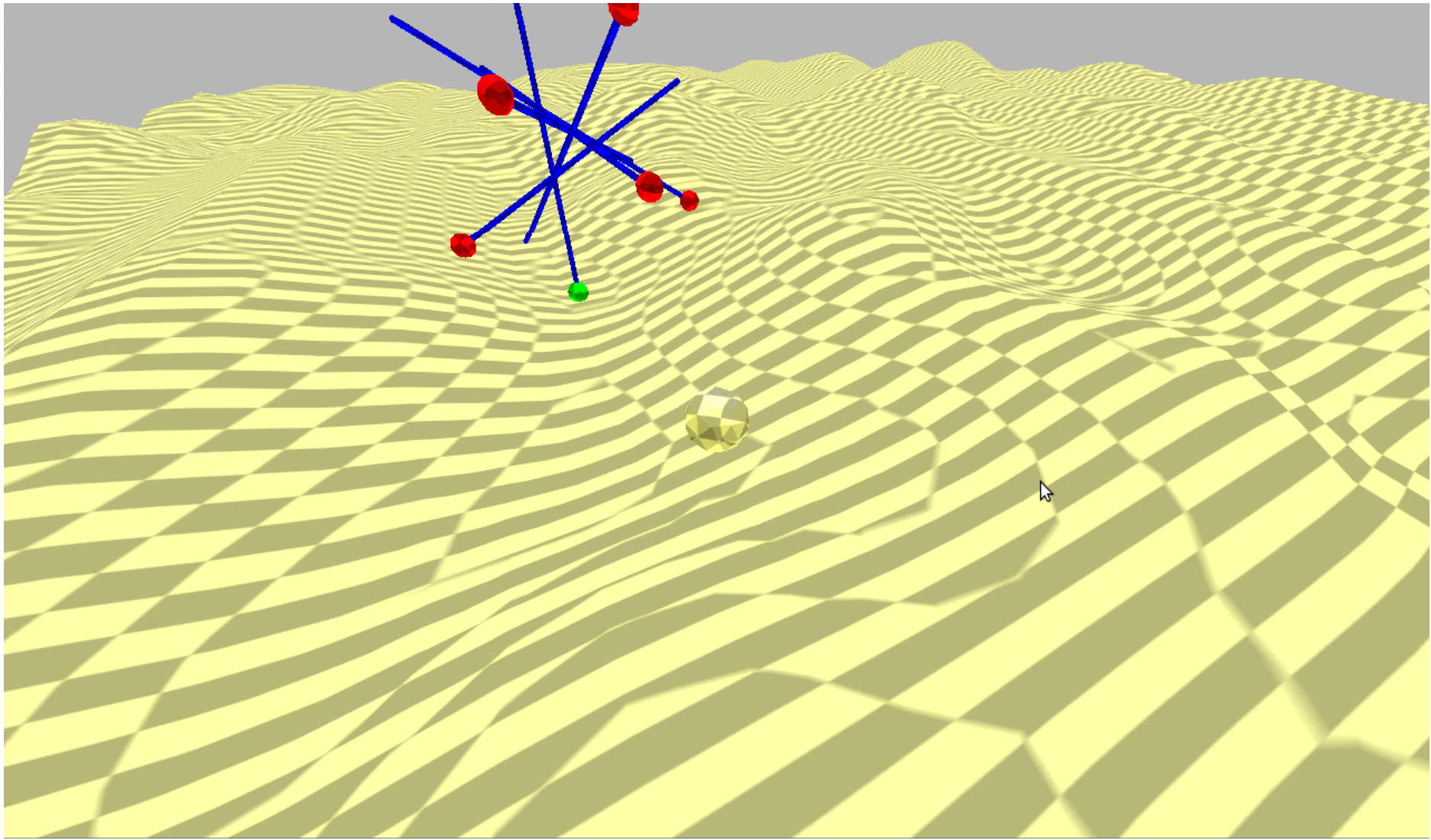}
\caption {Bullet Simulation - Non Planar Locomotion.}
\label{nonplanar}
            \end {center}
        \end {figure}

A physics engine is computer software that provides an approximate simulation of certain simple physical systems, such as rigid body dynamics (including collision detection). Their main uses are in mechanical design and video games. Bullet \cite{bullet} is an open source physics engine featuring 3D collision detection, soft body dynamics, and rigid body dynamics. The robot's structure was coded in about 60 Bullet code lines. Weights move up and down the legs using sliders (a slider is a Bullet object materializing the link between rigid bodies) as shown in Figure \ref{closeup}. To illustrate the robot's operation in real time, we added a target sphere to which the user can apply a force vector using the keyboard's $\leftarrow\rightarrow\uparrow\downarrow$ keys. As the target sphere starts to move, the robot starts revolving to follow it. We could hence visually conduct realistic physical experiments on various surfaces with the robot. {\sl cf.} Figures \ref{planar} and \ref{nonplanar}. A movie showing such an experiment is available on \cite{claret}.\smallskip

\medskip

\section{Further Research}

This work raises a number of interesting questions that seem to deserve attention:

\subsubsection{Landing State Probability:} Assume that the robot is given a random 3D spin and is thrown on a planar surface. What are the probabilities $\Pr_{\ell,L}[\mbox{{\sc hu}}], \Pr_{\ell,L}[\mbox{{\sc hd}}]$ and $\Pr_{\ell,L}[\mbox{{\sc sd}}]=1-\Pr_{\ell,L}[\mbox{{\sc hu}}]- \Pr_{\ell,L}[\mbox{{\sc hd}}]$ that the robot falls into each of the states? \smallskip

\subsubsection{Energy:} It is equally interesting to compute the energy spent during locomotion and finding out if for a given locomotion task there exists an optimal worm drive lifting strategy. Indeed, it might be the case that weights must not necessarily be lifted until the end of each hollow leg but to a lesser energy-optimal height.\smallskip

\subsubsection{Inertia:} Taking inertia into account is interesting as well: inertia allows to capitalize spent energy by keeping rolling instead of halting at each locomotion step. This is very apparent in the Bullet simulation but quite difficult to model precisely.\smallskip

\subsubsection{Slopes:} Finally, it is interesting to determine the robot's maximal climbable slope $\alpha_c(\ell,L,m)$ as well as the robot's maximal controlled descending slope $\alpha_a(\ell,L,m)$. A controlled descending is a descent of a slope in which the robot can halt at any point {\sl i.e.} not roll down a hill.\smallskip

Last but not least, it would be interesting to physically construct a working prototype of the device.


\begin{thebibliography}{99}

\bibitem{bullet} {\tt http://bulletphysics.org/}

\bibitem{claret} {\tt http://guillaume.claret.me/bunach/}

\bibitem{astar} P. Hart, N. Nilsson, B. Raphael, (1968). A Formal Basis for the Heuristic Determination of Minimum Cost Paths. IEEE Transactions on Systems Science and Cybernetics SSC4 4 (2): 100--107.

\bibitem{bfs} D. Knuth, (1997), The Art Of Computer Programming Vol 1. 3rd ed., Boston: Addison-Wesley.

\end{thebibliography}
\end{document}